\documentclass[10pt,twocolumn,letterpaper]{article}

\newcommand\blfootnote[1]{%
  \begingroup
  \renewcommand\thefootnote{}\footnote{#1}%
  \addtocounter{footnote}{-1}%
  \endgroup
}

\usepackage{cvpr}              %
\usepackage{float}
\usepackage{placeins}
\usepackage[table]{xcolor}
\usepackage{collcell}
\usepackage[moderate]{savetrees}
\usepackage{comment}

\newcommand{\SouS}[1]{#1}

\newcommand{\pg}[1]{#1}
\newcommand{\qheading}[1]{\noindent\textbf{#1}}

\newcommand{\ourmethod}[0]{RAVEN}
\newcommand{\stdnormal}[1]{\mathcal{N}(\mathbf{0}, I^{#1\times#1})}

\definecolor{cvprblue}{rgb}{0.21,0.49,0.74}
\usepackage[pagebackref,breaklinks,colorlinks,citecolor=cvprblue]{hyperref}

\def\paperID{78} %
\def\confName{3DV\xspace}
\def\confYear{2025\xspace}

\title{RAVEN: Rethinking Adversarial Video Generation \\ with Efficient Tri-plane Networks}

\author{Partha Ghosh$^{\dag 1}$,\enspace Soubhik Sanyal$^{\dag 1}$, \enspace Cordelia Schmid$^{*2}$, \enspace Bernhard Schölkopf$^{2\dag\ddag}$\\
$^\dag$ Max Planck Institute for intelligent Systems, Tübingen, Germany \\
$^*$ Inria, \'Ecole normale sup\'erieure, CNRS, PSL Research University\\
$^\ddag$ ELLIS Institute T\"ubingen, Germany\\
{\tt\small \{pghosh, soubhik.sanyal, bernhard.schoelkopf\}@tuebingen.mpg.de, cordelia.schmid@inria.fr}
}

\begin{document}

\twocolumn[{%
	\renewcommand\twocolumn[1][]{#1}%
	\maketitle
	\begin{center}
            \centerline{
                \includegraphics[width=0.31\linewidth]{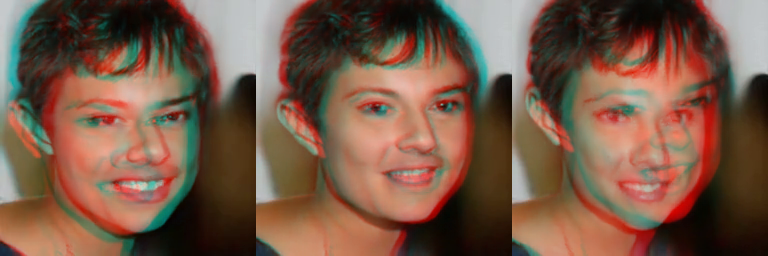}
                \includegraphics[width=0.31\linewidth]{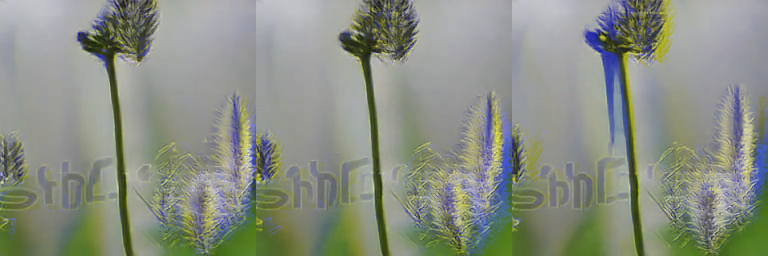}
                \includegraphics[width=0.31\linewidth]{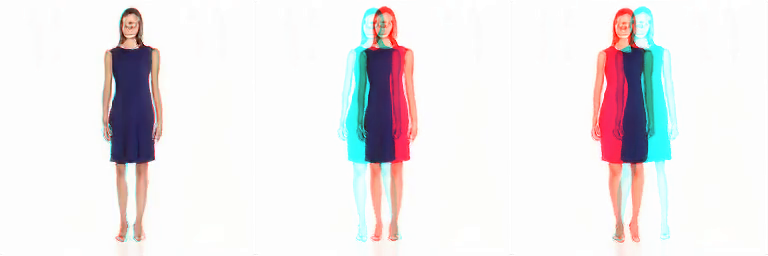}
            }
            \centerline{
                \includegraphics[width=0.31\linewidth]{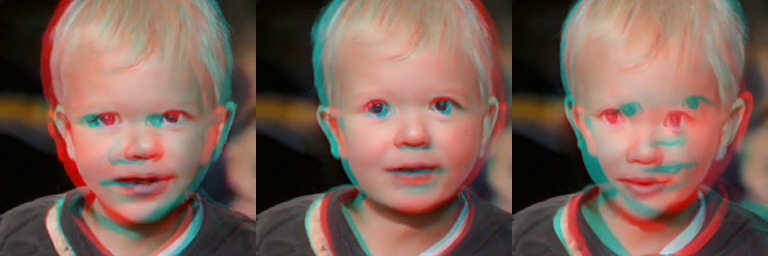}
                \includegraphics[width=0.31\linewidth]{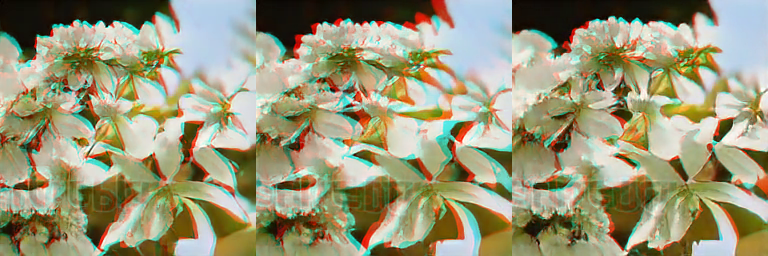}
                \includegraphics[width=0.31\linewidth]{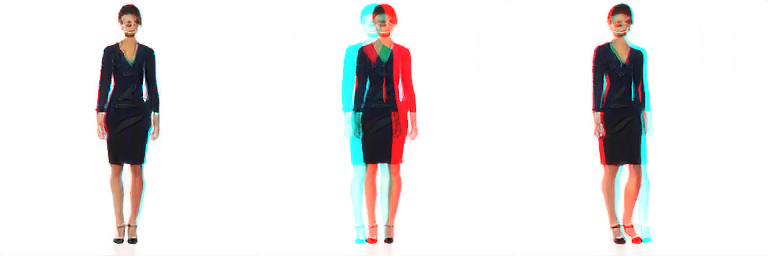}
            }
            \captionof{figure}{Here are samples of frames generated by our method, trained on three different datasets: Talking Faces, Webvid10M-Flowers, and Fashion videos. We select six consecutive frames spaced about 0.8 seconds apart to emphasize the motion between frames in the generated video. Then, we swap the red color channel between each pair of consecutive frames. Specifically, the red channel of the 0th frame is swapped with the 1st, the 2nd with the 3rd, and the 4th with the 5th. This results in three overlapping frames with visible motion across the color channels. The videos are available in the supplementary material.}
            \label{fig_teaser}
        \end{center}
}]

\maketitle

\blfootnote{$^{1, 2}$ Equal contribution.}

\begin{abstract}

We present a novel unconditional video generative model designed to address long-term spatial and temporal dependencies, \pg{with attention to computational and dataset efficiency}. To capture long spatio-temporal dependencies, our approach incorporates a hybrid explicit-implicit tri-plane representation inspired by 3D-aware generative frameworks developed for three-dimensional object representation and employs a single latent code to model an entire video clip. Individual video frames are then synthesized from an intermediate tri-plane representation, which itself is derived from the primary latent code. 
This novel strategy more than halves the computational complexity measured in FLOPs compared to the most efficient state-of-the-art methods.
Consequently, our approach facilitates the efficient and temporally coherent generation of videos. 
Moreover, our joint frame modeling approach, in contrast to autoregressive methods, mitigates the generation of visual artifacts. 
We further enhance the model's capabilities by integrating an optical flow-based module within our Generative Adversarial Network (GAN) based generator architecture, thereby compensating for the constraints imposed by a smaller generator size. 
As a result, our model synthesizes high-fidelity video clips at a resolution of $256\times256$ pixels, with durations extending to more than $5$ seconds at a frame rate of 30 fps. 
The efficacy and versatility of our approach are empirically validated through qualitative and quantitative assessments across three different datasets comprising both synthetic and real video clips. We will make our training and inference code public.

\end{abstract}
    
\section{Introduction}
\label{sec:intro}

The field of video generation has made remarkable progress in recent years. This development has especially focused on the generation of high-quality content-guided videos that are both realistic and efficient. This progress, however, has not reached the same heights as image generation due to two primary challenges: the inherent complexity of video data and the intensive computational demand of video processing. \pg{We choose a GAN-based approach over more performant diffusion-based ones keeping computational and dataset demands in mind.}

Most existing unconditional video generation methods leverage high-performing image generation models. To create videos, they generate one frame at a time autoregressively. This technique, while effective, is widely accepted to often accumulate error over the regression chain. This is also known as exposure bias~\cite{martinez2021pose, DU2023110996, li2023learning, ak2020incorporating}. Moreover, image generators are often imperfect, even when they are carefully trained with underlying 3D structures in narrow domain~\cite{GIF2020}. This further complicates building autoregressive models that use pre-trained image generators. Content-guided video generation, on the other hand, relies on additional modalities and is dependent on training with extensive datasets and complex models~\cite{xing2023make, esser2023structure}.

To overcome the limitations associated with autoregressive models in unconditional video generation, we propose a new representation method. Drawing inspiration from the tri-plane representation of 3D objects~\cite{chan2022efficient}, we adapt this approach to video data.

Our methodology begins by drawing parallels between the representation of 3D objects in neural networks and the challenges faced by video generative models. A 3D object, when modeled as a Neural Radiance Field (NeRF)~\cite{mildenhall2020nerf}, is described as a continuous volumetric scene function. 

By analogy, a video is a 3D array of color values indexed by two spatial coordinates and one temporal coordinate. Thus, representing a video with a parametric function involves mapping a 3D input (x, y, t) to RGB pixel values, effectively parameterizing a video as a continuous volume of color.

When designing a generative model with a continuous output space, various forms can be employed, such as explicit voxel grids~\cite{gadelha20173d, henzler2019escaping} or implicit representations~\cite{chan2021pi, niemeyer2021giraffe}, which are prevalent in 3D object modeling. However, these representations, while effective for single scenes, have proven unsuitable for high-resolution generative models due to their prohibitive memory requirements~\cite{chan2022efficient, wu2016learning, zhu2018visual}. Consequently, we adopt the tri-plane representation, which reduces memory demands significantly. There have also been studies that explore the effectiveness of multiplane representation of videos such as K-Planes~\cite{fridovich2023k}, HexPlane~\cite{cao2023hexplane}, and Tensor4D~\cite{shao2023tensor4d}. However, they do not extend them in the context of a generative model.

Despite the promise of the tri-plane representation for video, we have yet to fully understand its inductive biases, especially given the fundamental differences between spatial and temporal coordinates. Objects in space rarely appear identical at different locations, but when time is introduced as a coordinate, the same object often appears repeatedly, either stationary or moving. We hypothesize that to represent repeated instances of an object at different coordinates using a tri-plane, the model must feature redundancy in its representation. The neural network producing these features must therefore be capable of generating identical outputs for varied inputs, a challenging task without strategic design considerations.

To address this, we incorporate optical flow and warping operations, allowing the model to maintain feature consistency over time. This technique is corroborated by previous studies~\cite{ni2023conditional, 9481904} that successfully employed optical flow in the feature domain to capture motion dynamics.

In summary, our research presents three main contributions to the field of video generation:
1) We introduce a tri-plane representation for video data, adept at capturing long-range spatio-temporal dependencies.
2) We develop a generative model design that efficiently handles the creation of extended video sequences.
3) We present a novel optical flow-based motion model that enhances the representation of motion in our generative framework.

These enable us to produce high-quality, photorealistic videos at a resolution of 256x256 pixels for 160 frames at a frame rate of 30 fps. This marks a tenfold improvement in temporal upscaling and a sixteenfold enhancement in spatial resolution, all while requiring three times less computational effort compared to current state-of-the-art models. Moreover, our model supports test-time frame extrapolation and interpolation, which significantly advances the video generation process.

\section{Related work}
\label{sec:related_work}

\begin{figure*}
    \centering
    \includegraphics[width=0.95\textwidth]{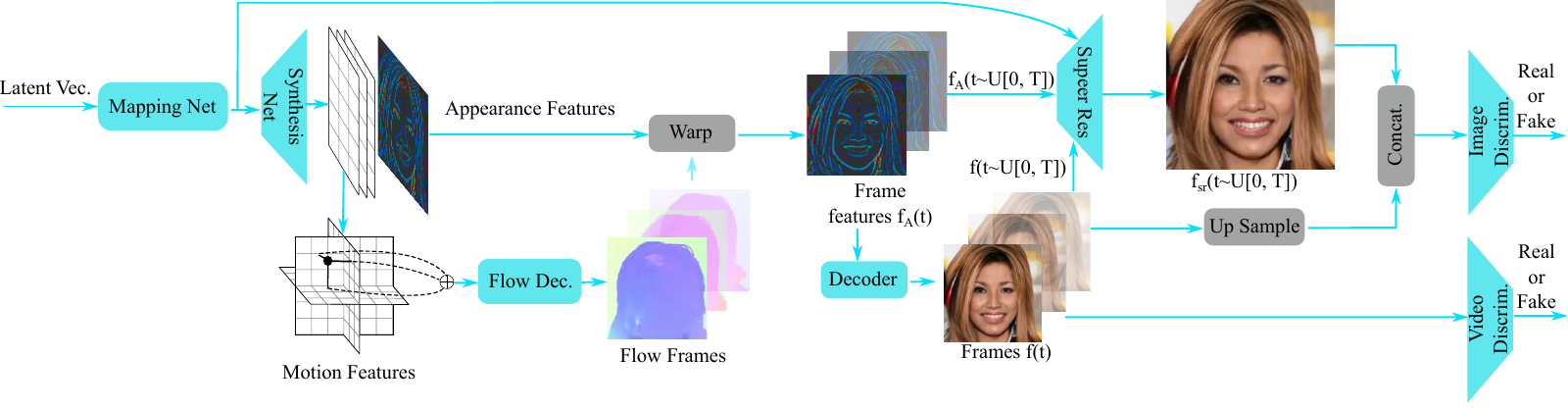}
    \caption{Our video generation model comprises the following parts: a StyleGAN-t-based backbone, a tri-plane representation of motion, a flow decoder, a forward warping process, a super-resolution module, an image discriminator, and a video discriminator. This architecture allows efficient representation of video data thanks to the tri-plane representation. It also represents motion explicitly, thanks to the flow fields and warping mechanism. Finally, we discriminate the generated video in low resolution and random frames in high resolution, allowing for the generation of efficient high-resolution video of long duration. Here blocks in cyan color represent trainable modules and blocks in gray color represent fixed operation.}
    \label{fig:our_model}
\end{figure*}

In the evolving landscape of video generation research, a spectrum of methodologies and frameworks has been proposed to address various complexities that arise. Broadly, the literature can be categorized into three core thematic domains: Content-Guided Video Generation, Computational Efficiency and Training Strategies, and Long-Term Temporal Dynamics. This overview aims to elucidate the contributions in each domain while emphasizing their strengths and synergies.

\qheading{Content-guided video generation:}
Content-guided video generation aims to generate videos in adherence to specific content instructions, often furnished through textual prompts or visual descriptors. Among these, the paper on Structure and Content-Guided Video Synthesis with Diffusion Models~\cite{esser2023structure} offers an innovative solution to the intricate challenge of segregating content and structural elements within a video, thereby granting granular control over individual video attributes. However, it requires a prerecorded video to borrow the content from at test time. Another noteworthy approach, dubbed Phenaki~\cite{villegas2023phenaki}, excels in open-domain video generation from textual prompts and stands out for its robust generalization abilities and spatio-temporal consistency. However, it still suffers from the non-rigid deformation of rigid objects and therefore lacks realism. Adding to this category, GODIVA~\cite{wu2021godiva} employs a pre-trained model to tackle the computational challenges inherent to text-to-video synthesis. However, it operates with a large number of correlated spatio-temporal tokens and suffers from similar shortcomings as Phenaki. GODIVA boasts fine-tuning capabilities and offers zero-shot performance on unseen textual descriptions. Recently released closed-source approaches like SORA~\cite{videoworldsimulators2024} further expand the scope of text-guided video generation, but little details are known.

\qheading{Computational efficiency and training strategies:}
Efficiency in computational resources and training paradigms constitutes another critical avenue of investigation. For instance, MagicVideo~\cite{zhou2022magicvideo} prioritizes computational efficiency by adopting a low-dimensional latent space combined with a 3D U-Net architecture, particularly focusing on high-resolution videos. StyleGAN-V~\cite{skorokhodov2022stylegan} presents a paradigm shift by conceptualizing video generation as a continuous-time signal, leveraging sparse training data to enhance computational efficiency. Additionally, Video Diffusion Models~\cite{ho2022video} extend the architecture proposed for image diffusion to facilitate video generation, aiming for temporal coherence and high-fidelity outputs. This model capitalizes on a joint training regimen utilizing both image and video data, thereby accelerating optimization processes and enhancing video quality. Albeit more efficient than competitors, these methods typically generate a video one frame at a time and the diffusion-based ones require many denoising steps per frame, making them computationally heavy.

\qheading{Long-Term temporal dynamics:}
Another subset of the research corpus addresses the inherent complexity of modeling long-term temporal dynamics in video sequences. The paper titled ``Generating Long Videos of Dynamic Scenes"~\cite{brooks2022generating} pioneers in this realm by focusing on the accurate reproduction of object trajectories, viewpoint alterations, and content changes over extended time frames. It innovatively refactors temporal latent variables and undergoes training on lengthier videos to circumvent limitations in contemporary methodologies. MoCoGAN~\cite{tulyakov2018mocogan}, on the other hand, proposes a comprehensive framework that separates video into distinct content and motion components, employing adversarial learning techniques along with specialized image and video discriminators to validate its efficacy. 
The methods discussed above generally follow the setting where temporal correlation among frames comes from the inductive bias that the latent codes from which individual frames are generated are correlated. This however suffers from two major drawbacks. The first is that of computational complexity. This is because the whole model has to run one time for every frame. Second, since every frame is related to its neighbors only via the latent code, it has to account for both fine-grained temporal consistency such as simple pixel-wise translation, and at the same time object-level consistency such as articulated movement or rotation. Since these consistency constraints are not spread across different resolutions of intermediate features, we see non-rigid deformations and unrealistic artifacts sip into the results.

In summary, our work addresses the aforementioned problems by introducing a tri-plane representation of video data, which efficiently models long-range spatiotemporal correlations. We also propose an efficient generator architecture specifically designed to handle extended video sequences. Moreover, we introduce an optical flow-based motion model to facilitate the creation of unconditional, photorealistic videos at a resolution of \(256 \times 256\) pixels spanning $160$ frames. This design further accommodates test-time frame extrapolation and interpolation capabilities. Finally, since a whole video clip is generated in one shot, the intra-frame correlation is enforced automatically.

\section{Method}
\label{sec:method}

\ourmethod~is an unconditional model for generating videos, featuring a generator $\mathcal{G}$ that outputs a complete 160-frame video clip $\mathcal{V}$ from a single input noise vector $z$ drawn from a 512-dimensional standard normal distribution, denoted as $z\sim\stdnormal{512}$. Formally, we express this as $\mathcal{V} = \mathcal{G}(z)$. The generator $\mathcal{G}$ is equipped with a backbone network responsible for producing tri-plane representation features and includes a flow module to enhance the generator's capacity to represent motion. We use two separate discriminators which together with the generator are trained using the adversarial objective.

\subsection{Tri-plane hybrid video representation}
\label{subsec:triplane_fpr_vid}

Generating high-resolution videos necessitates an efficient data representation due to the impractical computational demand of direct application of 3D convolutions. In this section, we adapt the tri-plane representation from EG3D~\cite{chan2022efficient} for video data, a technique initially proposed to efficiently represent 3D objects. A suitable representation of 3D objects for generative modeling faces similar memory complexities as ours. Videos, like 3D objects, can be considered three-dimensional arrays, with videos being arrays of pixels and objects as arrays of material properties (density and radiance properties).

By overfitting the tri-plane representation to a single video, we establish its suitability for video data. This involves organizing features into three planar grids aligned with the spatial and temporal axes —-- $xt$, $yt$, and $xy$, with $x$, $y$ as the spatial axes, and $t$ as the time axis. Each grid has dimensions $N \times N \times C$, where $N$ is the spatio-temporal resolution and $C$ is the number of channels. To determine a pixel's color at any location in space and time, we project the point onto the three planes, obtain the features $F_{xt}, F_{yt}, F_{xy}$ at the projection point by bilinear interpolation, and then decode them using a lightweight decoder network composed of two StyleGAN synthesis blocks.

This tri-plane approach is more memory-efficient than a full 3D voxel-based video representation, as it can represent a whole video with only three 2D arrays of features. It is also more memory efficient than other implicit video representations, as it uses a simple decoder network and effectively uses the representation capacity of the feature planes. To confirm the efficacy of this representation for videos, we assess its performance through frame interpolation~(\cref{table:quantitative_comaprison_interpolation}) and extrapolation~(\cref{table:quantitative_comaprison_extrapolation}) experiments, using SSIM and PSNR as our quality metrics, and comparing it against both fully implicit~\cite{barron2021mip} and dense feature volume representation of equivalent memory usage. Results show that the tri-plane representation, especially when augmented with a flow module, outperforms other forms of representation across various datasets and tasks.

\begin{table*}[h!]
\centering
    \begin{tabular}{lcccccccccccc}
        \toprule
        & \multicolumn{4}{c}{Talking Faces all Motions} & \multicolumn{4}{c}{Fashion Videos} & \multicolumn{4}{c}{WebVid10M-Flowers} \\
        \cmidrule(lr){2-5} \cmidrule(lr){6-9} \cmidrule(lr){10-13}
        Method     & \multicolumn{2}{c}{3 frame interp.} & \multicolumn{2}{c}{8 frame intrp.}& \multicolumn{2}{c}{3 frame interp.} & \multicolumn{2}{c}{8 frame intrp.}& \multicolumn{2}{c}{3 frame interp.} & \multicolumn{2}{c}{8 frame intrp.} \\ 
        \cmidrule(lr){2-3}\cmidrule(lr){4-5}\cmidrule(lr){6-7}\cmidrule(lr){8-9}\cmidrule(lr){10-11}\cmidrule(lr){12-13}
         & SSIM & PSNR & SSIM & PSNR & SSIM & PSNR & SSIM & PSNR & SSIM & PSNR & SSIM & PSNR \\
        \midrule
        Pos. emb & 0.14 &14.43 &0.15 &14.34 &0.66 &15.54 &0.67 &15.60 &0.21 &15.56 &0.20 &15.46  \\
        voxels & 0.75 &26.29 &0.74 &26.08 &0.89 &26.62 &0.88 &26.09 &0.59 & 22.41 &0.59 & 22.48  \\
        triplane  & 0.88 &31.19 &0.87 &30.48 &0.95 & \textbf{32.38} &0.93 &30.10 & \textbf{0.71} &24.98 &\textbf{0.71} & 24.84  \\
        triplane+flow & \textbf{0.92} & \textbf{32.77} & \textbf{0.90} & \textbf{32.08} & \textbf{0.96} & 32.25 & \textbf{0.95} & \textbf{30.71} & \textbf{0.71} & \textbf{25.33} & 0.70 & \textbf{24.94}  \\
        \bottomrule
    \end{tabular}
    \caption{\textbf{Tri-plane video representation for interpolation:} We compare the effectiveness of the tri-plane representation for Video data by computing SSIM and PSNR value of interpolated frames. Note that for all metric, higher is better. The reported numbers here are average PSNR over $20$ random videos taken from each of our data sets}
    \label{table:quantitative_comaprison_interpolation}
\end{table*}

\begin{table*}[h!]
\centering
    \begin{tabular}{lcccccccccccc}
        \toprule
        & \multicolumn{4}{c}{Talking Faces all Motions} & \multicolumn{4}{c}{Fashion Videos} & \multicolumn{4}{c}{WebVid10M-Flowers} \\
        \cmidrule(lr){2-5} \cmidrule(lr){6-9} \cmidrule(lr){10-13}
        Method     & \multicolumn{2}{c}{3 frame interp.} & \multicolumn{2}{c}{8 frame intrp.}& \multicolumn{2}{c}{3 frame interp.} & \multicolumn{2}{c}{8 frame intrp.}& \multicolumn{2}{c}{3 frame interp.} & \multicolumn{2}{c}{8 frame intrp.} \\ 
        \cmidrule(lr){2-3}\cmidrule(lr){4-5}\cmidrule(lr){6-7}\cmidrule(lr){8-9}\cmidrule(lr){10-11}\cmidrule(lr){12-13}
         & SSIM & PSNR & SSIM & PSNR & SSIM & PSNR & SSIM & PSNR & SSIM & PSNR & SSIM & PSNR \\
        \midrule
        Pos. emb & 0.17 &15.26 &0.17 &15.19 &0.70 &17.41 &0.70 &17.59 & 0.29 & 17.51 & 0.29 & 17.4  \\
        voxels & 0.73 &25.71 &0.71 &24.77 &0.86 &24.87 &0.85 &23.95 & 0.49 & 21.72 & 0.49 & 21.82 \\
        triplane  & 0.83 &28.49 &0.82 &27.61 &0.91 &27.65 &\textbf{0.90} &26.40 & \textbf{0.58} & 23.49 & \textbf{0.57} & 23.40  \\
        triplane+flow & \textbf{0.86} &\textbf{29.76} &\textbf{0.84} &\textbf{28.32} &\textbf{0.92} &\textbf{28.25} &\textbf{0.90} &\textbf{26.65} &0.57&\textbf{23.57} &\textbf{0.57} &\textbf{23.53}  \\
        \bottomrule
    \end{tabular}
    \caption{\textbf{Tri-plane video representation for extrapolation:} We compare the effectiveness of the tri-plane representation for Video data by computing SSIM and PSNR value of extrapolated frames. Note that for all metric, higher is better. The reported numbers here are average PSNR over $20$ random videos taken from each of our data sets}
    \label{table:quantitative_comaprison_extrapolation}
\end{table*}

\subsection{Tri-Plane + Flow: Explicit motion modeling}
\label{subsec:triplane_plus_flow}

To generate realistic videos, it is essential to account for object translations while maintaining appearance consistency. In a tri-plane representation of a video with object translations (and rotations), feature replication across the planes is inevitable. Given that these features originate from a generative model (in our case a GAN generator), the model must learn to replicate identical features accurately. It is a non-trivial task and its accuracy can't be guaranteed. To circumvent this, our approach first generates global and local flow fields from the tri-plane features. Then the initial grid of features is warped and alpha-matted. This effectively manages translations, occlusions, and background changes.

We begin by generating the tri-plane feature grids $F_{tri}$ and a separate global feature plane $F_G$, derived from the output of the backbone StyleGAN network. At each point in a $64\times64\times64$ grid, we sample features from the tri-plane grid, referred to as motion features. These motion features are then decoded by a two-layer MLP (labeled as `Flow decoder' in~\cref{fig:our_model}, $f_{fd}:\mathbb{R}^{36} \rightarrow \mathbb{R}^5$) into, local and global flow vectors and a masking scalar (together represented as `Flow frames in~\cref{fig:our_model}').

We obtain the time-dependent feature frame $F_r(t)$ as follows. First, using the global flow field $f^G_t$ at a given time $t$, we warp the global feature frame $F_G$ to obtain the time-dependent global features $F_r^G(t)=w(F_G, f^G_t)$. Here $w$ is the forward warping function. Concurrently, the local flow $f^L_t$ helps adjust the preceding feature frame $F_r(t-1)$ of time step $(t-1)$ to obtain the time-dependent local features $F_r^L(t)=w(F_r(t), f^l_t)$. These local and global features are alpha-blended to obtain the time-dependent feature at time $t$ as $F_r(t) = m_t*F_r^G(t) + (1-m_t)*F_r^L(t)$. The mask $m_t$ is obtained, as $m_t = f_{fd}(F_{tri}, t)$, is the output of the flow decoder $f_{fd}$. Mask $m_t$ is an additional output alongside the flow fields. By sequentially performing this process for all time steps, we create a $64\times64\times64$ grid of appearance features. We call this grid as the appearance volume. Somewhat similar to our method LEO~\cite{wang2023leo}, also relies on optical flow. However, it does so in the image space. Our method on the other hand performs flow-based warping in the latent space. This approach results in only 64 forward warping steps of a $64\times64$ feature grid to generate a video with $256$ frames. This, compared to $256$ warping steps on a larger grid required by LEO for the same video clip size, is significantly more computationally efficient. Additionally, by operating in the feature space, our method is able to represent complex 3D motions and address occlusion and background dynamics, which are often challenging for traditional optical flow based generation methods.

To determine a pixel's color, we interpolate a feature from the appearance volume at the spatio-temporal location of the pixel and decode it with a network comprising two StyleGAN synthesis blocks (labeled simply as `Decoder' in~\cref{fig:our_model}). The incorporation of flow fields into our model shows marked improvements in frame interpolation~(\cref{table:quantitative_comaprison_interpolation}) and extrapolation~(\cref{table:quantitative_comaprison_extrapolation}) tasks.

\subsection{CNN backbone of Tri-plane and generation} 

The tri-plane's composition of three 2D feature arrays allows for the use of a conventional 2D CNN generator. We selected the StyleGAN-T architecture for its exceptional capacity for feature representation. Adapting StyleGAN-T's output, our model generates a $64\times64\times68$ feature grid, partitioned into a 32-channel global feature grid ($F_G$) and three $12$-channel motion feature grids. From these, we derive local and global flow fields and a mask, which are utilized to warp and blend the appearance features, forming an appearance feature volume.
To farther enhance training efficiency, we sample a frame at a random time $t$ at a $64\times64$ resolution and generate the entire video at $32\times32\times32$. We achieve the final $256\times256$ resolution through image-domain super-resolution, applied only to the selected frame to conserve computational resources. This process mirrors the approach taken in eg3D, utilizing two StyleGAN2 synthesis blocks to upscale the image by a factor of four.

\subsection{Double Discrimination} 

Similar to a standard GAN setting, we use a discriminator to criticize the generated data. We deploy two different discriminators, one for the frame of time step $t$, and the other for the generated video. We extend the built-in StyleGAN-T discriminator to accommodate $6$ channels instead of $3$. Such that it can take both the low and high-resolution versions of the generated frame and force the super-resolution module to maintain consistency. Specifically, we take the ViT-based feature extractor networks of StyleGAN-t networks as is and run them two times, once on the low-resolution image and once on the high-resolution version of it. The feature extractors work at a constant resolution, we therefore adjust the resolution as needed with bilinear interpolation. Finally, the extracted features are concatenated along the channel axis and are fed to the trainable discriminator heads of StyleGAN-T. The generated video is passed through a video discriminator that has the same design as the video discriminator described in \cite{brooks2022generating}.

\subsection{Implementation details}

We utilize the most extensive version of the StyleGAN-T generator, which contains around one billion parameters, and apply its progressive training approach. Training starts with the generator producing $16 \times 16$ resolution for both appearance and motion feature planes, and gradually, we increase this to $128 \times 128$. The video output resolution remains unchanged during training, irrespective of the feature plane resolution. To match the increasing resolution of the feature planes, we progressively reduce the Gaussian blur applied to the frames of the dataset, starting from a sigma value of 10 and halving it each time the feature resolution is increased, following the sequence: $10,$ $5,$ $2.5,$ $1.25,$ $0$. The transition to a higher resolution is triggered when the FID score ceases to improve for randomly sampled frames. The full training process takes about $5$ days on $8$ Nvidia A100 GPUs, with a consistent batch size of $128$ throughout the training sessions with Adam Optimizer. We found that larger batch sizes have the effect of stabilizing training. For all evaluation purposes, we use the same learning rate for the generator($2.5\times10^{-3}$) and the discriminator($2\times10^{-3}$) and Adam optimizer with $\beta_1 = 0$ and $\beta_2=0.99$. Following standard StyleGAN-T training, we use the R1 penalty for the discriminators. We use a loss weight ratio of $3:1$ between the video to image discriminator. 

\section{Experiments}
\label{sec:experiments}

We evaluate \ourmethod~on the video generation properties~(\cref{fig_teaser}), compare it to the state-of-the-art methods on synthetic and real datasets, and perform a detailed set of ablation studies. 
Here we primarily focus on comparing our method to other GAN-based methods, keeping in mind apples-to-apples comparison. We however acknowledge the importance of application-focused comparisons, \ie comparison against all works that generate videos, such as~\cite{hoppe2022diffusion,hong2023cogvideo,Zhou2022MagicVideoEV}\etc. We separate this in~\cref{sec:diffusion_comparison}.

\subsection{Datasets}
\label{subsec:datasets}

Our model was evaluated across various datasets featuring distinct motion types, specifically focusing on 1) Talking faces with all motions and a subset with 10 specific motions, 2) Fashion videos, and 3) WebVid10M-Flowers dataset, all at a resolution of \(256 \times 256\). Similar to image-based GANs, we observed that video GANs perform better with larger datasets. Due to the insufficiency of existing datasets in terms of size, we introduced the first two datasets. 
\SouS{Additionally, we collected videos of flowers moving with the wind to form a subset of the WebVid10M~\cite{Bain21} dataset to evaluate our model on an existing real-world dataset. According to the download page, the original WebVid10M dataset is no longer available.}

\qheading{Talking Faces:} Public datasets such as CelebV-HQ~\cite{zhu2022celebv} and Face-Forensics~\cite{roessler2018faceforensics} are inadequate in size and quality for our needs. We generated a synthetic dataset of approximately 400 thousand videos of talking faces using Thin-plate spline motion models~\cite{zhao2022thin} to animate FFHQ faces~\cite{karras2019style} with CelebV-HQ motions~\cite{zhu2022celebv}. However, this process yielded many low-motion videos, prompting us to curate a subset of 10 high-motion videos from CelebV-HQ~\cite{zhu2022celebv} to create a more dynamic Talking Faces dataset.

\qheading{Fashion Videos:} To capture a variety of motions beyond facial movements, we created the Fashion Videos dataset. The original dataset from Dwnet~\cite{zablotskaia2019dwnet} is too small, so we synthesize additional data by combining motions from pairs of videos. The Thin-plate spline~\cite{zhao2022thin} and First order motion models~\cite{NEURIPS2019_31c0b36a} proved inadequate for this task, leading us to adopt the Bidirectional Deformable Motion Modulation method~\cite{yu2023bidirectionally}. We further refined the dataset by excluding videos with long dresses that caused inconsistencies, resulting in a total of 37,285 videos.

\pg{\qheading{WebVid10M-Flowers:} To form the WebVid10M-Flowers dataset we collect a subset of videos from WebVid10M~\cite{Bain21} dataset that contain flowers moving in the wind. We perform this using similarity matching of the captions of the videos. We use the sentence transformer~\cite{reimers-2019-sentence-bert} to perform sentence-level embedding and retrieval. The query that we used is ``flowers moving in the wind".}

\subsection{Comparisons} 
Following video generation research norms, we use Fréchet Video Distance (FVD)~\cite{unterthiner2018towards} and Fréchet Inception Distance (FID)~\cite{heusel2017gans} as our main evaluation metrics. Consistent with the approaches of Skorokhodov et al.~\cite{skorokhodov2022stylegan}, we utilize two variants of FVD, namely, $\text{FVD}_{16}$ and $\text{FVD}_{128}$—which correspond to the analysis of video clips containing 16 and 128 frames, respectively. For FID assessment, we compare 50,000 randomly generated frames against 50,000 original frames sampled from the dataset. We adopt the FVD computation method from StyleGAN-V~\cite{skorokhodov2022stylegan}. Notably, we opt for mp4 format to store generated and real videos to minimize memory usage. For instance, storing 2048 videos with 160 frames each consumes only 60 MB in mp4 format, in stark contrast to the 32 GB required for uncompressed png frames. This efficiency alleviates the memory constraints that previously limited StyleGAN-V~\cite{skorokhodov2022stylegan} to using only 2048 samples for metric estimation. To maintain comparability with existing benchmarks, we continue to use 2048 samples for metric evaluation. Our method is benchmarked against the leading GAN-based approaches in unconditional video generation, namely StyleGAN-V~\cite{skorokhodov2022stylegan} and MoCoGAN~\cite{tulyakov2018mocogan} with a StyleGAN2 backbone. \pg{We also compare with the state-of-the-art open source diffusion-based video generation method `stable video diffusion (SVD)'~\cite{blattmann2023stable}. We note that strictly speaking it is not a fair comparison as the model is image conditioned \ie, it takes as input the first frame. Furthermore, to compare under the same dataset we finetune a pre-trained SVD model on our dataset. Therefore it has an overwhelming data advantage over any of the GAN-based models. We still include these results in our quantitative evaluation section. }

\begin{figure*}[t]
\centering

        \begin{subfigure}{0.49\linewidth}
          \centering
              \includegraphics[width=1.01\linewidth]{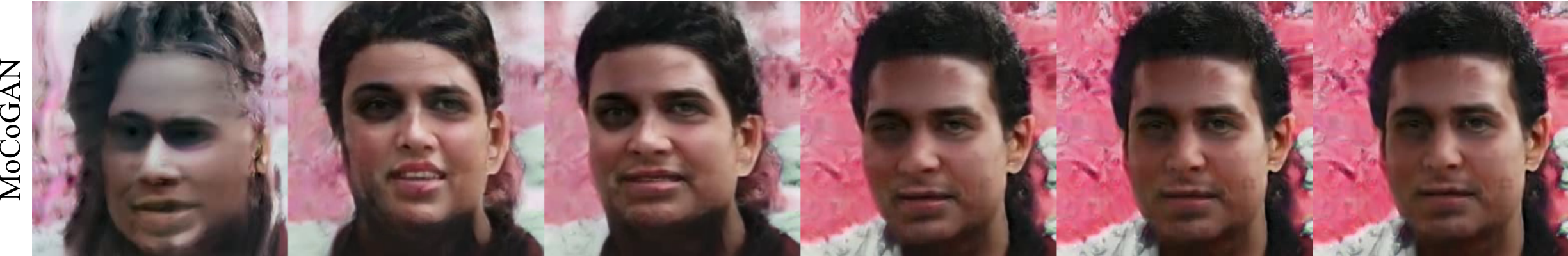}
              \includegraphics[width=1.01\linewidth]{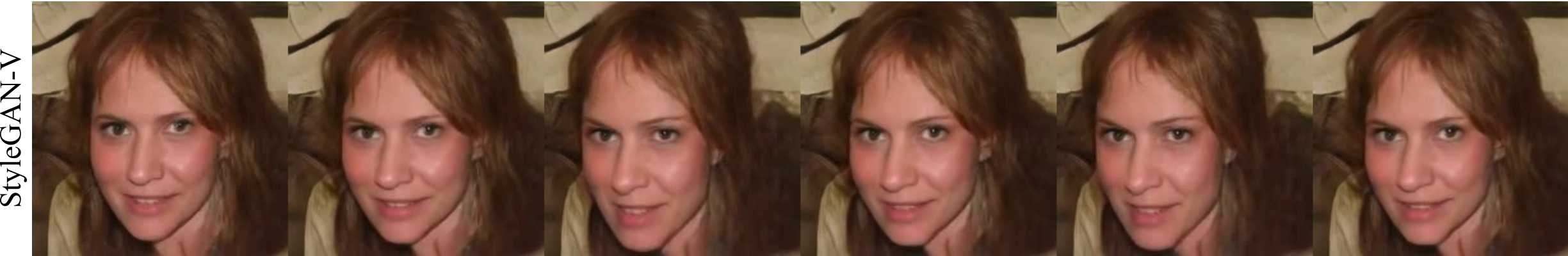}
              \includegraphics[width=1.01\linewidth]{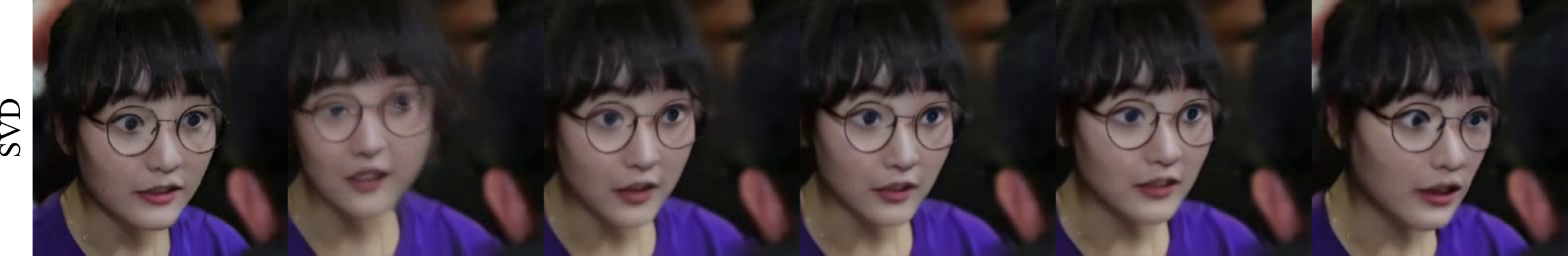}
              \includegraphics[width=1.01\linewidth]{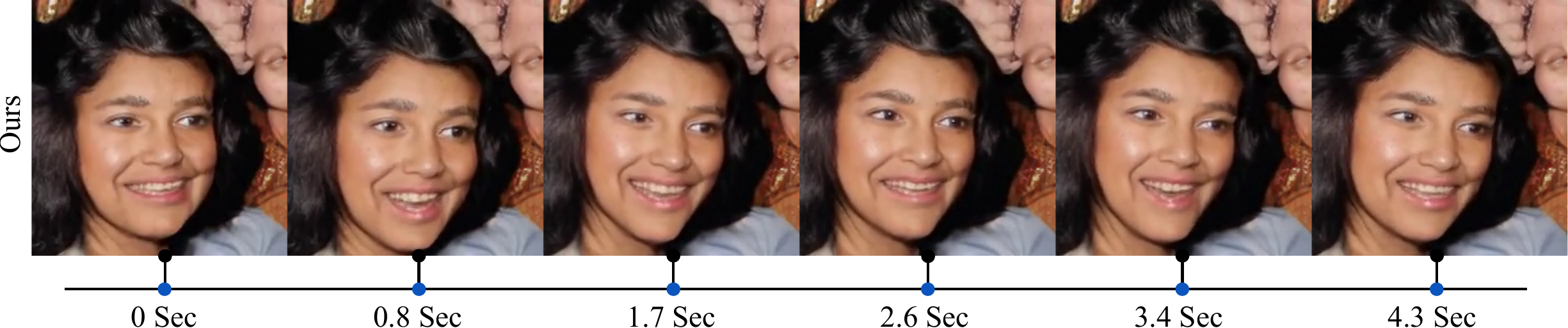}
    
          \caption{}
          \label{Fig:qualitative_results_a}
        \end{subfigure}      
        \begin{subfigure}{0.49\linewidth}
          \centering
            \includegraphics[width=0.967\linewidth, trim=20mm 0 0 0, clip]{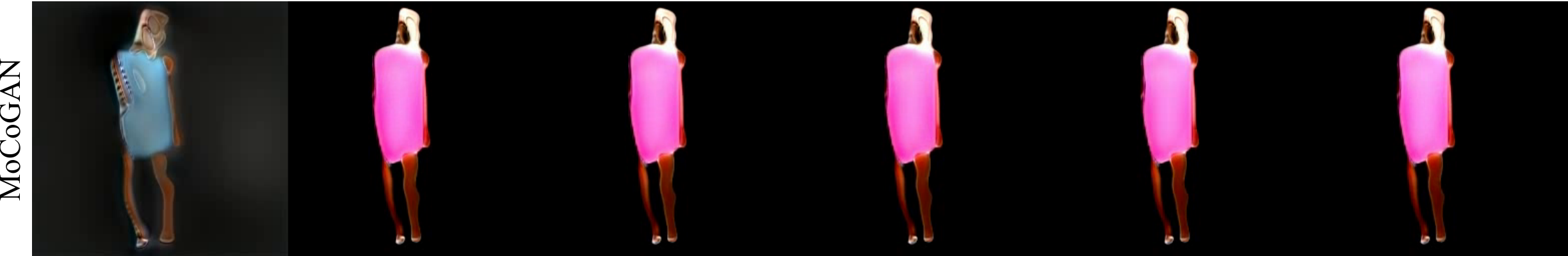}
            \includegraphics[width=0.967\linewidth, trim=20mm 0 0 0, clip]{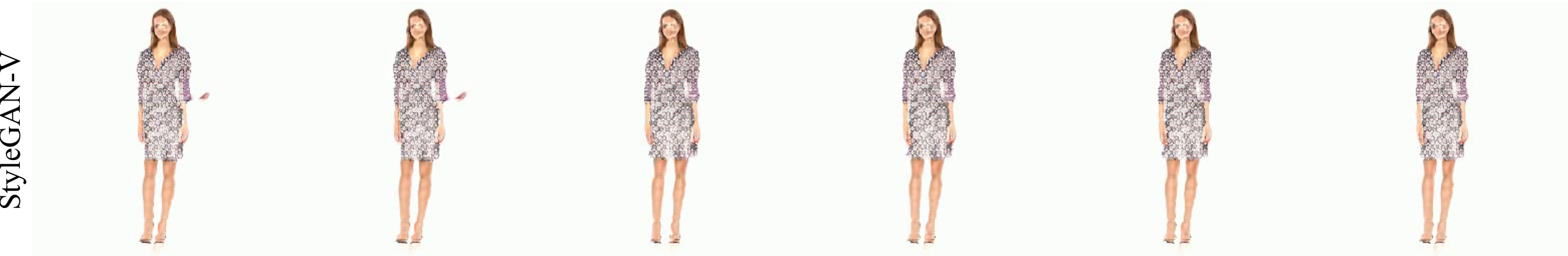}
            \includegraphics[width=0.967\linewidth, trim=20mm 0 0 0, clip]{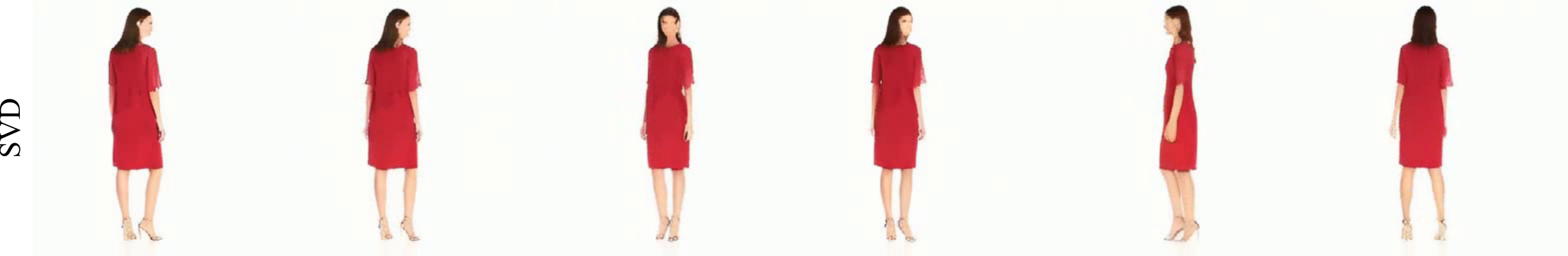}
            \includegraphics[width=0.967\linewidth, trim=20mm 0 0 0, clip]{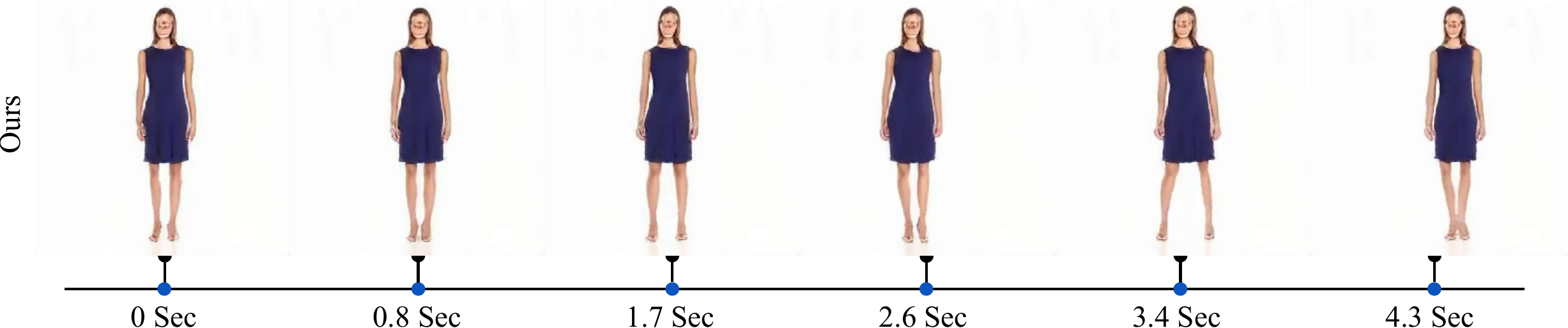}
          \caption{}
          \label{Fig:qualitative_results_b}
    \end{subfigure}
    
    \caption{\textbf{Qualitative results:} Here in~\cref{Fig:qualitative_results_a}~and~\cref{Fig:qualitative_results_b} we show the qualitative results generated by our method in comparison to MoCoGAN~\cite{tulyakov2018mocogan}, StyleGAN-V~\cite{skorokhodov2022stylegan} and stable video diffusion (SVD)~\cite{blattmann2023stable}. For visualization, we only show $6$ equally spaced frames from a video clip of length $5$ seconds \pg{for all models except for SVD. Since SVD only generates 25 frames, we simply visualize 6 equally spaced frames (\ie we skip 4 frames in between every shown frame for SVD).} \pg{We provide several generated videos from each model in the supplementary material.}}
\end{figure*}

\qheading{Qualitative evaluation:}
In~\cref{Fig:qualitative_results_a}~and~\ref{Fig:qualitative_results_b} \ourmethod~is qualitatively compared with StyleGAN-V~\cite{skorokhodov2022stylegan} and MoCoGAN~\cite{tulyakov2018mocogan} with enhanced backbones \pg{as well as stable video diffusion (SVD)~\cite{blattmann2023stable}. Here we examine videos with a total duration of $5.3$ seconds at $30$ fps by selecting every $21$st frame for display. However, since SVD only generates 25 frames, we simply display 6 equally spaced frames.} The complete videos are provided in the supplementary materials in mp4 format. \pg{In ~(\cref{fig_teaser}) by overlapping different color channels, we highlight the movement of the objects generated by our method. Moreover,} our approach demonstrates its ability to capture the long-range spatial and temporal dependencies in the data, as shown by the consistency in the identity of subjects in talking face sequences and the coherent changes in appearance and limb movement in fashion videos. Notably, in the talking-faces dataset, StyleGAN-V shows a similar performance to our method, which could be due to two factors. Firstly, the model capacity required for representing talking faces, particularly with minimal motion, is well within StyleGAN2's capabilities. Secondly, the dataset we synthesized from CelebV-HQ and FFHQ, using Thin-plate spline motion models, inherently features limited motion, particularly due to the prevalence of smiling expressions with partially open mouths in FFHQ, resulting in constrained jaw movement. These factors diminish the performance distinction between our method and StyleGAN-V. However, for the Talking Faces 10 motions dataset, which includes videos with more pronounced motions, our method shows a greater performance margin as indicated in the quantitative comparison table~(\cref{table:quantitative_comaprison}). The superiority of our approach becomes more apparent in the fashion video dataset, which presents more complex structures such as articulated human figures, where other state-of-the-art methods struggle significantly.

\paragraph{Quantitative evaluation}

\begin{table*}[h]
\centering
\setlength{\tabcolsep}{5pt}
    \begin{tabular}{lcccccccccc}
        \toprule
        & \multicolumn{3}{c}{Talking Faces all/$10$ Motions} & \multicolumn{3}{c}{Fashion Videos} & \multicolumn{3}{c}{WebVid10M-Flowers} & \# T-FLOPs \\
        \cmidrule(lr){2-4} \cmidrule(lr){5-7} \cmidrule(lr){8-10}
        Method  & FID & FVD16 & FVD128 & FID & FVD16 & FVD128 & FID & FVD16 & FVD128\\
        \midrule
        StyleGAN-V & 11.8/18.6 & 135/135 & 112/138 & 19.66 &167.3& 167.9 & 32.30 & 142.2 &  -- & 10.125\\
        StyleGNA-V-T & 166.5/-- & 257.6 & 2851 & 302.4 & 1259 & 1532 & 303.7 & 744.6 & -- & 43.039\\
        MOCOGAN & 58.89/-- & 209.4/-- & 348.2/-- & 202.9 & 686.6 & 1700 & 38.64 & 145.9 & -- & 10.092\\
         \midrule
        Ours-Stg2 & 323.3/-- & 4605/-- & 2781/-- & 298.3 & 967.0 & 2114 & FT & FT &  FT & \textbf{2.688}\\
        Ours-StgXL & 26.49/-- & 116.4/-- & 203.0/-- & 26.28 & 72.46 & 219.2 & FT & FT &  FT & 7.686\\
        Ours-StgT & \textbf{8.9}/\textbf{4.7} & \textbf{97}/\textbf{65.6} & \textbf{110.3}/\textbf{62.1} & \textbf{10.8} & \textbf{67.5} & \textbf{111.1} & \textbf{8.71} & \textbf{119} & 190 & 4.183\\
        
        \bottomrule
    \end{tabular}
    \caption{We compare the performance of our method against state-of-the-art diffusion-based and GAN-based methods under three different datasets and two different metrics, FID and FVD. We use $2048$ video samples, $16$, and $128$ consecutive frames to estimate FVD. To estimate FID, we use real random frames from all the videos available and 50K generated random video frames. FT: fails to train.}
    \label{table:quantitative_comaprison}
\end{table*}

Our evaluation, detailed in the \cref{table:quantitative_comaprison}, indicates that our video generation method outperforms current leading GAN-based alternatives\footnote{We forego comparison with MOCOGAN-HD with its original backbone as its performance is comprehensively beaten by StyleGAN-V, and we outperform StyleGNA-V on all datasets under all metrics. We instead compare to the enhanced version of MOCOGAN proposed in StyleGAN-V.}. This success is attributed to our method's ability to train across the entire video volume, unlike other state-of-the-art (SOTA) approaches. We investigate if StyleGAN-V's performance can be improved by simply changing the generative backbone to a stronger model as compared to StyleGAN2, namely StyleGAN-T. We call this method StyleGAN-V-T. However, as noted in~\cite{sauer2023stylegan}, we also observe that simply changing the generative backbone is not enough and in fact it degrades its performance. Our model also demonstrates a significant reduction in computational effort to generate a single video sample of $160$ frames, requiring less than half the FLOPs compared to other SOTA models. However, our use of a per-frame super-resolution network, while parallelizable and rapid (running at 54 FPS on a Quadro RTX 6000 GPU), it does increase the total computational expense, accounting for about 50\% of our compute budget at test time.

\pg{Furtheremore, we compare under the same metric the performance of SVD~\cite{blattmann2023stable} against our method. We fine-tune a pre-trained SVD model on our datasets for this comparison. However, as mentioned earlier there remains a lot of inherent nonuniformity in training procedures simply because the two methods are widely different in their scope. First, SVD generates only 25 frames as compared to 160 frames that our method generates. This immediately poses the challenge of framerate selection for the training dataset. We subsample our training dataset by 5 fold to preserve a similar amount of motion complexity for SVD. Second, SVD uses the first frame as a condition, therefore FID computation on the first frame becomes noninformative. We therefore do not report this metric. Third, the FVD2048\_128f metric requires 128 frames in every sample. This is unavailable for SVD, we therefore report only FVD2048\_16f. The performance of SVD under the FVD2048\_16f metric are $87.7$, $260.9$, and $196.6$ for the three of our datasets talking-faces, fashion videos, WebVid10M-Flowers respectively. Please refer to the Supmat for qualitative results. As can be seen, \ourmethod~is competitive to SVD despite SVD having been pretrained on data that is many-fold larger than ours.}

Our method's ability to one-shot generate all frames for inspection by the discriminator contrasts sharply with SOTA methods that generate frames sequentially from weakly correlated latent codes. This capability, along with our model's efficient tri-plane representation, circumvents the need for any additional post-processing to achieve the desired video resolution for discriminator evaluation. By employing separate discriminators for motion and appearance, we simplify the evaluation process for each discriminator, allowing for a more straightforward assessment of long-range spatiotemporal correlations.

While our primary comparisons are with the most recent GAN-based video generation models, there is a wide range of research employing non-GAN architectures, such as autoregressive~\cite{yan2021videogpt, NEURIPS2022_94461854, ge2022long} and diffusion-based~\cite{esser2023structure, Blattmann_2023_CVPR, Yang2022DiffusionPM, Ge_2023_ICCV} models. Direct comparisons to these models are often not feasible due to the significant computational resources and time required to retrain and fine-tune these models on our datasets. Additionally, since these models are based on different generative mechanisms and often utilize distinct cropping, resizing, and metric evaluation methods, comparisons must be approached with caution.

Even so, we provide an FVD-based comparison of these architectures~(\cref{table:FVD_in_ucf}), with a note that such comparisons are not perfectly aligned. We also present the parameter counts for these models (taken from Wang et al.~\cite{wang2023videofactory}), noting that diffusion models, despite their parameter efficiency, require multiple iterations to refine their outputs, substantially increasing their effective computational load.

Our findings show that while our model achieves competitive results at a resolution of $256\times256$ with a relatively good FID score, the task of generating high-resolution videos remains a challenge. Some models exhibit strengths at lower resolutions but experience performance decline at higher resolutions. This highlights the dynamic nature of video generation research, where each approach has its advantages and limitations, and where there is a continuous search for methods that balance performance across various resolutions and complexities.

\newcommand{\graycell}[1]{\ifgray\expandafter\textcolor{gray}{#1}\else#1\fi}

\newif\ifgray
\grayfalse %

\newcolumntype{G}{>{\collectcell\graycell}c<{\endcollectcell}}
\newcolumntype{L}{>{\collectcell\graycell}l<{\endcollectcell}}

\begin{table}[h]
\setlength{\tabcolsep}{2pt}
    \centering
    \setlength{\tabcolsep}{3.3pt}
    \begin{tabular}{L@{\hskip -0.05mm}GGGGG}
        \toprule
        Method  & FID & \multicolumn{2}{c}{FVD} & Res. & \#param\\
        & & 16f & 128f & & \\
        \midrule \vspace{-12pt}
        \global\graytrue \\
        CogVideo~\cite{hong2023cogvideo}$^*$ & -- & 701.5 & -- & 160X160$^\dag$ & 9B \\
        Make-A-vid~\cite{Singer2022MakeAVideoTG}$^*$ & -- & 367.2 & -- & 64X64$^\dag$ & 7.36B$^\ddagger$\\
        MagicVideo~\cite{Zhou2022MagicVideoEV}$^*$ & 145 & 655.0 & -- & 256X256$^\dag$ & -- \\
        Video LDM~\cite{Blattmann_2023_CVPR}$^*$  & -- & 550.6 & -- & 128X256$^\dag$ & 912M$^\ddagger$ \\
        VideoFactory~\cite{wang2023videofactory}$^*$  & -- & 410.0 & -- & 344X192$^\dag$ & 1.6B\\
        PYoCo~\cite{Ge_2023_ICCV}$^*$ & -- & 355.1 & -- & 64X64$^\dag$ & 1.3B$^\ddagger$\\
        Vid. Diffusion~\cite{NEURIPS2022_39235c56} & 295 & -- & -- & 64X64 & 1.2B \\
        \vspace{-12pt}
        \global\grayfalse \\
        DIGAN~\cite{yu2022digan} & -- & 1630 & 2293 & 256X256 & --\\
        StyleGAN-V~\cite{skorokhodov2022stylegan} & -- & 1431 & 1773 & 256X256 & 32M\\
        MOCOGAN~\cite{skorokhodov2022stylegan} & -- & 1630 &  2293 & 256X256 & 29M\\
        Ours & 79.74 & 1136 & 1933 & 256X256 & 1B\\
        
        \bottomrule
    \end{tabular}
    \caption{We summarize the performance of recent video generative models from diverse backgrounds and design choices. All these metrics are reported on UCF101 dataset. In the rightmost column (titled Res.), we report the train resolution as reported metrics are resolution-sensitive. Note that $*$-marked methods are text-conditioned video generation. Methods marked $\dag$ train additional spatial and temporal super-resolution methods (sometimes text conditioned and as many as $4$ stages) to offer a higher-resolution inference model. For a fair comparison, we gray out those methods that operate in significantly different settings.%
    }
    \label{table:FVD_in_ucf}
\end{table}

\subsection{Ablation Studies}

Our principal contribution is the innovative use of a tri-plane representation for video data, aligning the planes with the spatial and temporal dimensions of the video. Our empirical validation demonstrates that tri-plane representation outperforms other efficient implicit, explicit, or hybrid forms with comparable memory usage, in terms of both extrapolation and interpolation capabilities across various datasets and window sizes, as detailed in our extrapolation and interpolation comparison tables~~\cref{table:quantitative_comaprison_extrapolation,table:quantitative_comaprison_interpolation}.

We also identify that the dynamic nature of video content, which frequently involves the translation of objects or of entire scenes, can lead to redundant feature replication that could otherwise be utilized for enhancing representational capacity. To address this, we introduce the novel concept of feature flow within our model, designed to optimize the utilization of representational capacity by minimizing redundancy. The effectiveness of this design is evident in the improved performance metrics~(\cref{table:quantitative_comaprison_extrapolation,table:quantitative_comaprison_interpolation}).

In addition, we employ the most expressive version of StyleGAN currently available, which we posit as the most suitable choice for our framework. This assertion is substantiated by a comparative analysis against a lesser model employing StyleGAN2's default settings. The comparative results, reflected in the rows labeled Ours-Stg2, Ours-StgXL, and Ours-Full in~\cref{table:quantitative_comaprison_interpolation}, clearly demonstrate the advantage of utilizing the larger StyleGAN model in our approach. In Ours-Stg2, Ours-StgXL, and Ours-Full we have used StyleGAN-2, StyleGAN-XL, and StyleGAN-t as the backbone of our generator respectively.

\subsection{Comparisons to Non-GAN-based approaches}
\label{sec:diffusion_comparison}
 We summarize the performance of recent video generative models from diverse backgrounds and design choices in~\cref{table:FVD_in_ucf}. We report FVDs as are found in the published papers of the indicated methods. There are variations in the training and evaluation protocol, such as crop size, normalization scheme, compression used to store original data, feature extraction network used \etc as also noted in~\cite{skorokhodov2022stylegan}. This makes drawing a quick conclusion a hard task as the precise FVD number is sensitive to the factors mentioned above. We however do think that one obtains useful insight from this comparison nonetheless. Here we would like to highlight, as an example, the comparison against Vidoe-LDM~\cite{Blattmann_2023_CVPR}. It achieves an FVD16 of $550.6$ in UCF 101 dataset at a resolution $128\times256$ which is $2\times$ lower than us. Despite being a diffusion-based model as opposed to our GAN based model, it uses a similar number of model parameters. However, it requires several stages of denoising and therefore significantly more floating point operations.

\section{Discussion}
\label{sec:discussion}
The effectiveness of \ourmethod~depends greatly on the generative backend. Consequently, the same design yields varied results, influenced by the network's capacity and dataset complexity. For instance, we've noted that the base model of StyleGAN2 performs reasonably well on simpler datasets. However, it significantly struggles with complex datasets like UCF101 or WebVid10M-Flowers. We believe that the current limitation in generation quality is primarily due to the capacity of the underlying backbone network.

In our observations, we've seen a consistent improvement in performance as we transitioned from the StyleGAN2 backbone to StyleGAN-XL and then to \mbox{StyleGAN-T}. This suggests that using a higher-capacity backbone, such as a diffusion model, may be an attractive option. However, diffusion models in particular come with their own set of challenges, particularly encoding video into the triplane representation, which is a nontrivial task. We leave it for future exploration.

Additionally, since we employ feature flow, akin to optical flow, to model foreground and background movements separately and in a disentangled manner, we inherit some complexities of optical flow. For example, in optical flow, the lack of explicit depth information necessitates modeling an additional mask to occlude the correct part of an object when two objects converge in the image plane. This can be addressed by using four planes instead of three, and thereby operating in 3 spatial dimensions, where occlusion can be handled more naturally.

Furthermore, the flow formulation encounters challenges when there is a substantial amount of unidirectional flow, leading to edge effects. While this can be mitigated by employing infinitely large feature planes, it would require a non-uniform grid structure to manage the memory requirements effectively.

Finally, it is important to highlight that \ourmethod currently does not provide direct and disentangled access to the objects within a scene it generates. This is a matter to be addressed in future research. The concept of disentanglement holds promise for video generative models, and we eagerly anticipate further investigations that leverage our approach to unlock its potential. Apart from the above, we also think a valuable direction to explore would be the in-painting and inversion capability of \ourmethod.

\section{Conclusion}
\label{sec:conclusion}
We present \ourmethod, a generative video model trained in an adversarial setting. We achieve memory and computational efficiency by carefully designing the generator and discriminator architecture that does not need to use expensive 3D convolution operation on high resolution. We achieve this by adapting the tri-plane representation first proposed in the 3D-aware generative modeling community. This helps us strike a balance between implicit and explicit representation of high dimensional data efficiently while capturing their complexity. We extend the modeling capacity of such representation using an explicit optical flow-like mechanism on the feature-level representation of a video. We evaluate our method extensively on many diverse data sets and showcase strong performance, using both quantitative and qualitative methods.

\paragraph{Acnowledgements}
This work was supported by the VideoPredict project, FKZ: 01IS21088.

\FloatBarrier
{
    \small
    \bibliographystyle{ieeenat_fullname}
    \bibliography{main}
}

\end{document}